\documentclass[letterpaper, 10 pt, conference]{ieeeconf}  

\usepackage{balance} 
\usepackage{amssymb}
\usepackage{tikz}
\usepackage{tikzducks}
\usepackage{amsmath}
\usepackage{graphicx}
\usepackage{booktabs}
\usepackage{amssymb}
\usepackage{pifont}
\usepackage{multirow}
\usepackage{float}
\usepackage{acro}
\usepackage{subcaption} 
\usepackage{caption}    
\usepackage{tabularx}    
\usepackage{makecell}
\usepackage{tabularx,booktabs,makecell,multirow}
\usepackage{siunitx}
\usepackage{caption}
\usepackage{hyperref}
\usepackage{cleveref} 

\newcolumntype{Y}{>{\raggedright\arraybackslash}X}
\usepackage{makecell} 
\newcommand{\cdash}{\makecell[c]{---}} 
\newcommand{\Yes}{\checkmark}
\newcommand{\YesN}{\checkmark\textsuperscript{*}}
\newcommand{\Dash}{---}
\newcommand{\TableSize}{\scriptsize}    
\newcommand{\TableStretch}{1.03}
\newcommand{\TableColsep}{2.2pt}
\newcommand{\UseUnifiedTableStyle}{%
  \TableSize
  \setlength{\tabcolsep}{\TableColsep}%
  \renewcommand{\arraystretch}{\TableStretch}%
}

\DeclareAcronym{RL}{short=RL,long=Reinforcement Learning}
\DeclareAcronym{DRL}{short=DRL,long=Distributional Reinforcement Learning}
\DeclareAcronym{IL}{short=IL,long=Imitation Learning}
\DeclareAcronym{DPPO}{short=DPPO,long=Distributional Proximal Policy Optimisation}
\DeclareAcronym{CVaR}{short=CVaR,long=Conditional Value at Risk}

\IEEEoverridecommandlockouts                              

\title{\LARGE \bf
Risk-Aware Reinforcement Learning for Mobile Manipulation
}

\author{Michael Groom$^{1*}$, James Wilson$^{2}$, Nick Hawes$^{1}$ and Lars Kunze$^{1,3}$
\thanks{$^{1}$Oxford Robotics Institute, Department of Engineering Science, University of Oxford, Oxford, United Kingdom}
\thanks{${^2}$Dyson Institute of Engineering \& Technology, Malmesbury, United Kingdom}
\thanks{$^{3}$Bristol Robotics Laboratory, School of Engineering, University of West England, Bristol, United Kingdom}
\thanks{$^{*}$Corresponding author: michaelgroom@robots.ox.ac.uk}
}

\begin{document}

\maketitle

\begin{tikzpicture}[remember picture,overlay]
\node[anchor=north, yshift=0pt] at (current page.north) {
    \parbox{\textwidth}{
        \centering \scriptsize \sffamily
        This work has been submitted to the IEEE for possible publication. \\
        Copyright may be transferred without notice, after which this version may no longer be accessible.
    }
};
\end{tikzpicture}

\thispagestyle{empty}
\pagestyle{empty}

\begin{abstract}
For robots to successfully transition from lab settings to everyday environments, they must begin to reason about the risks associated with their actions and make informed, risk-aware decisions. This is particularly true for robots performing mobile manipulation tasks, which involve both interacting with and navigating within dynamic, unstructured spaces. However, existing whole-body controllers for mobile manipulators typically lack explicit mechanisms for risk-sensitive decision-making under uncertainty. 
To our knowledge, we are the first to (i) learn risk-aware visuomotor policies for mobile manipulation conditioned on egocentric depth observations with runtime-adjustable risk sensitivity, and (ii) show risk-aware behaviours can be transferred through \ac{IL} to a visuomotor policy conditioned on egocentric depth observations.    
Our method achieves this by first training a privileged teacher policy using \ac{DRL}, with a risk-neutral distributional critic. Distortion risk-metrics are then applied to the critic's predicted return distribution to calculate risk-adjusted advantage estimates used in policy updates to achieve a range of risk-aware behaviours. We then distil teacher policies with \ac{IL} to obtain risk-aware student policies conditioned on egocentric depth observations. 
We perform extensive evaluations demonstrating that our trained visuomotor policies exhibit risk-aware behaviour (specifically achieving better worst-case performance) while performing reactive whole-body motions in unmapped environments, leveraging live depth observations for perception.  

\end{abstract}

\vspace{-0.3cm}
\section{Introduction}

Mobile manipulators combine a mobile base with a robotic arm to offer expansive workspaces for complex, open-ended tasks. However, these systems face compounded uncertainties from noisy localization, perception, and actuation. This aleatoric uncertainty over action execution and environment dynamics introduces significant risk regarding the value of potential outcomes.
Unlike structured settings with bounded uncertainty, dynamic, human-shared environments make standard expected-return optimisation unsafe; explicitly minimising severe outcomes is critical. 
By reasoning over a value distribution of potential outcomes rather than just their expected values, a risk-aware controller can assign greater weight to avoiding low-probability, high-cost catastrophic failures, enabling the robot to act commensurately with an acceptable level of risk.

\begin{figure}[!tp] 
    \centering 
    \includegraphics[trim={0cm, 1.7cm, 0cm, 0cm}, clip, width=\columnwidth]{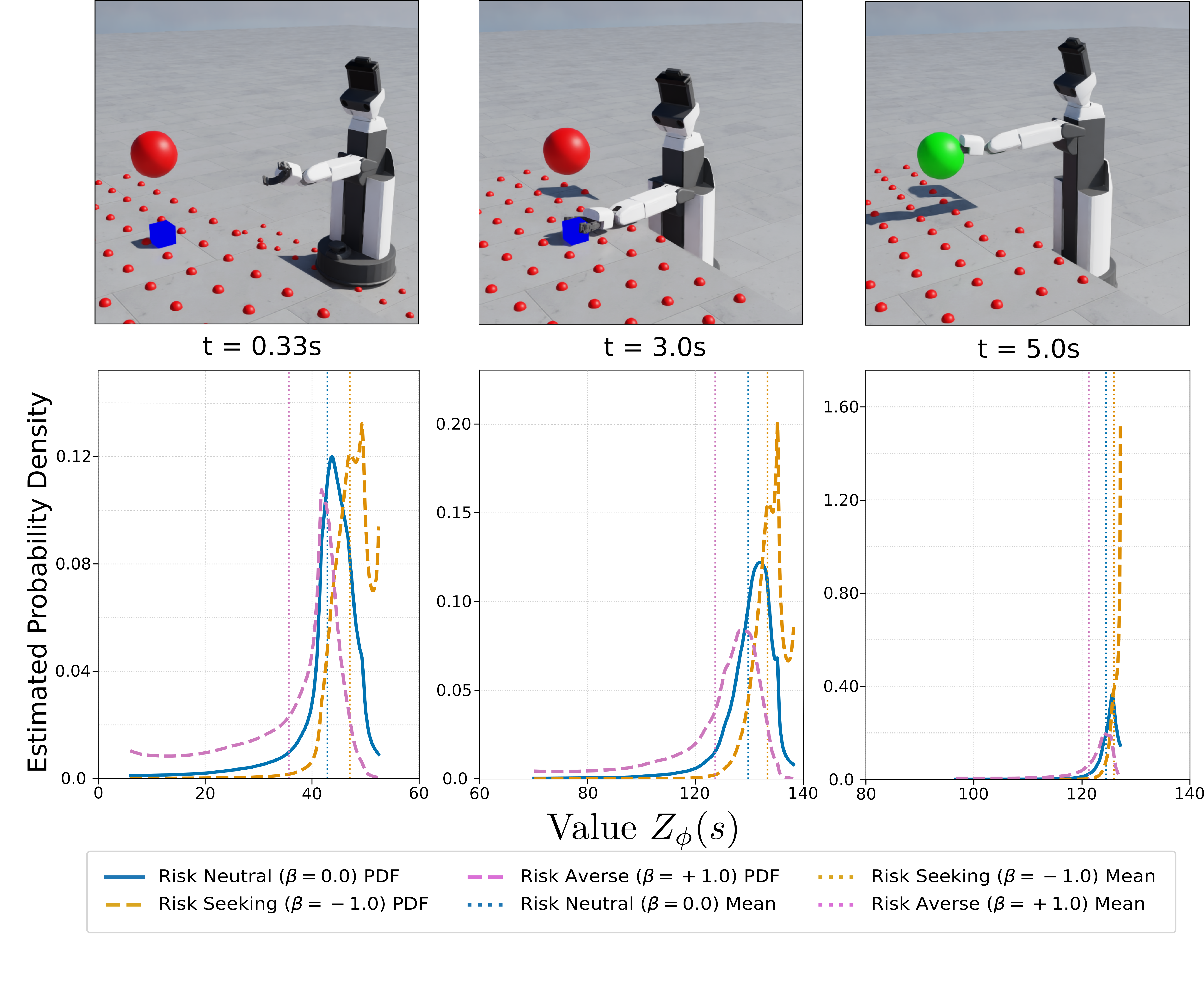} 
    \caption{Predicted value distributions from a risk-aware critic during a successful pick attempt at key time steps (PDFs via critic-predicted quantiles) shown for different risk attitudes: risk-neutral (\(\beta=0.0\)), risk-seeking (\(\beta=-1.0\)), and risk-averse (\(\beta=+1.0\)). Risk sensitivity alters the perceived relative likelihood of outcomes, producing risk-aware behaviour. Note changing axis ranges.
    }
    \label{fig:first_figure} 
    \vspace{-0.5cm}
\end{figure}

\begin{figure*}[t]
  \centering
  \includegraphics[trim={20pt 6pt 32pt 5pt}, clip, width=\textwidth]{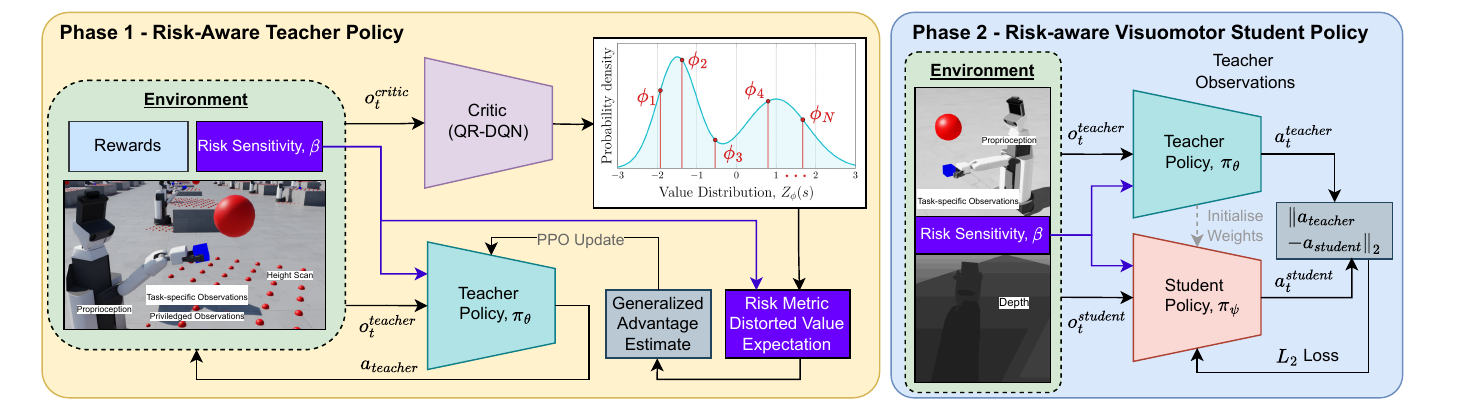}
  \caption{An overview of our proposed framework. \textbf{Phase 1:} A risk-aware \ac{DRL} teacher policy \(\pi_{\theta}\) is trained. A critic predicts value distributions \(Z_{\phi}(s)\) which are distorted by a risk metric to calculate a risk distorted expected value, which is used to update the teacher policy \(\pi_{\theta}\), which is also conditioned on the selected risk-sensitivity. \textbf{Phase 2:} A student policy \(\pi_{\psi}\) conditioned on high-dimensional depth observations is learnt through \ac{IL} with the risk-aware teacher policy \(\pi_{\theta}\). The risk-sensitivity parameter $\beta$ is assumed to be provided by an external operator or planner at runtime.} 
  \label{fig:method_diagram}
  \vspace{-0.45cm}
\end{figure*}

While traditional model-based motion planning approaches excel in static scenarios, they rely on rigid assumptions—such as perfect perception and up-to-date global maps—making real-time replanning computationally prohibitive in noisy, dynamic environments \cite{brunke2022safe, sandakalum2022motion}. Learning-based approaches, particularly \ac{RL} \cite{sutton2018reinforcement}, offer a compelling alternative by enabling reactive control directly from high-dimensional sensor data. However, standard \ac{RL} is typically formulated to maximize expected return, inherently ignoring the variance and tail-end risks of catastrophic outcomes \cite{majumdar2020distortion, bdr2023}. 

To allow explicit reasoning about risk, Distributional \ac{RL} models the full value distribution \cite{dabney2018distributional}, yet its application to training visuomotor policies remains largely unexplored due to severe sample inefficiency of online \ac{RL}. Teacher-student \ac{IL} \cite{ross2011reduction, cheng2024extreme} is a proven method for mitigating this inefficiency by distilling expert \ac{RL} policies into vision-based student networks. However, prior distillation frameworks have successfully distilled task-specific expertise without exploring the transfer of risk-aware policy behaviours.

To bridge this gap, we propose the first risk-aware \ac{RL} framework for mobile manipulation capable of runtime-adjustable risk sensitivity from raw sensory input. We first train an expert teacher policy using \ac{DPPO} \cite{Schneider-ICRA-24} on low-dimensional privileged states. Crucially, rather than estimating environmental risk directly, this teacher is conditioned on a dynamic risk-sensitivity parameter that allows a user or higher-level system to continuously modulate behaviour from risk-averse to risk-seeking. Using a teacher-student \ac{IL} framework \cite{ross2011reduction}, we then distil this risk-aware teacher into a student policy conditioned on non-privileged observations, including egocentric depth images. This allows the mobile manipulator to reactively perform whole-body tasks in unstructured, dynamic environments without a prior map, while maintaining the risk-sensitive behaviours learned in simulation.

In summary, our contributions are as follows:
\begin{enumerate}
    \item We introduce the first framework to combine \ac{DRL} with distortion risk metrics to train egocentric depth-based mobile manipulation policies, featuring a runtime-adjustable risk-sensitivity parameter.
    \item We demonstrate that learned risk-aware behaviours can be transferred via \ac{IL} to a vision-based student policy capable of whole-body control in unstructured, dynamic environments. 
\end{enumerate}

We support these contributions with a thorough analysis of policies trained with our framework, detailing emergent risk-aware behaviours.

\vspace{-0.2cm}
\section{Related Works}
\label{sec:related_works}

\textbf{Whole-body Mobile Manipulation.} While classical planners generate optimal trajectories \cite{ratliff_chomp_2009, kavraki_probabilistic_1996}, high replanning costs limit their reactivity in dynamic environments \cite{sandakalum2022motion}. Similarly, recent holistic controllers utilizing quadratic programming still rely on static maps and manual behaviour trees \cite{haviland_holistic_2022}. Conversely, learning-based frameworks (\textit{e.g.}, \ac{RL} and \ac{IL}) bypass explicit mapping, achieving notable success in reactive, sensor-driven whole-body control \cite{tang2025deep, fu2024mobile, liu2025visual, black2025pi0}. However, standard learning frameworks fail to explicitly reason about uncertainty-driven risk, restricting their safety-critical application \cite{brunke2022safe}. While constrained \ac{RL} addresses safety by maximizing expected returns subject to hard constraints \cite{dadiotis2025dynamic}, it often lacks conditioning on high-dimensional data and still ignores risk from uncertainty. To overcome this, risk-aware \ac{RL} applies soft constraints via distortion risk metrics, ensuring low-probability, low-return outcomes commensurately influence policy updates.
\\
\textbf{Risk-Aware RL and Distributional Methods.}
Standard \ac{RL} maximizes expected return, effectively ignoring the variance and tail-ends of potential outcomes \cite{sutton2018reinforcement}. Risk-aware \ac{RL} instead optimizes a risk-adjusted metric, such as \ac{CVaR} \cite{rockafellar2000optimization}. To avoid the time-inconsistency inherent in static risk evaluation, dynamic risk approaches apply risk metrics recursively \cite{majumdar2020distortion}. This is typically implemented via \ac{DRL} \cite{dabney2018distributional, ma2020dsac}.  Unlike standard \ac{RL} which predicts a scalar expected value, \ac{DRL} models the full return distribution, allowing risk metrics to be applied directly to the predicted outcomes.
In robotics, dynamic risk-aware \ac{RL} has been successfully applied to quadruped \cite{Schneider-ICRA-24, shi2024robust} and humanoid locomotion \cite{wu2025arc}, often dynamically adjusting risk sensitivity based on uncertainty estimations \cite{schubert2021automatic}. However, these methods rely heavily on low-dimensional proprioceptive states. While some distributional visuomotor policies exist \cite{bodnar2019quantile}, they are limited to static manipulators and rely on scripted exploration strategies. Consequently, there remains a critical gap in extending dynamic, risk-aware \ac{RL} to high-dimensional perception in mobile manipulation tasks.
\\
\textbf{Imitation Learning and Distillation.}
Applying \ac{RL} directly to high-dimensional visual observations in simulation is notoriously sample-inefficient \cite{uppal_spin_2024}. To overcome this in robotics, expert distillation paradigms are frequently employed: an expert ``teacher" policy 
(often privileged, state-based \ac{RL} policies \cite{cheng2024extreme, kumar2021rma}, or human teleoperation \cite{zhao2023learning}) is distilled into a ``student" policy conditioned on realistic, high-dimensional sensor data. While this \ac{IL} architecture effectively bridges the gap between state-based simulation and vision-based deployment, existing teacher-student frameworks for mobile manipulation lack mechanisms for risk-awareness. Our work addresses this limitation by integrating dynamic risk-aware \ac{RL} into a distillation pipeline, enabling risk-aware, reactive mobile manipulation driven by raw sensory input.

\section{Method}
\label{sec:method}



We present a two-phase framework for risk-aware mobile manipulation: training a privileged risk-aware teacher policy (Sec. \ref{Sec: method: teacher policy}), followed by distillation into a vision-based student policy via imitation learning (Sec. \ref{Sec: method: student policy}). 
\subsection{Risk-aware Privileged Teacher Policy (Phase 1)}
\label{Sec: method: teacher policy}
\begin{figure}[htbp]
  \centering
  \includegraphics[width=\columnwidth]  
  {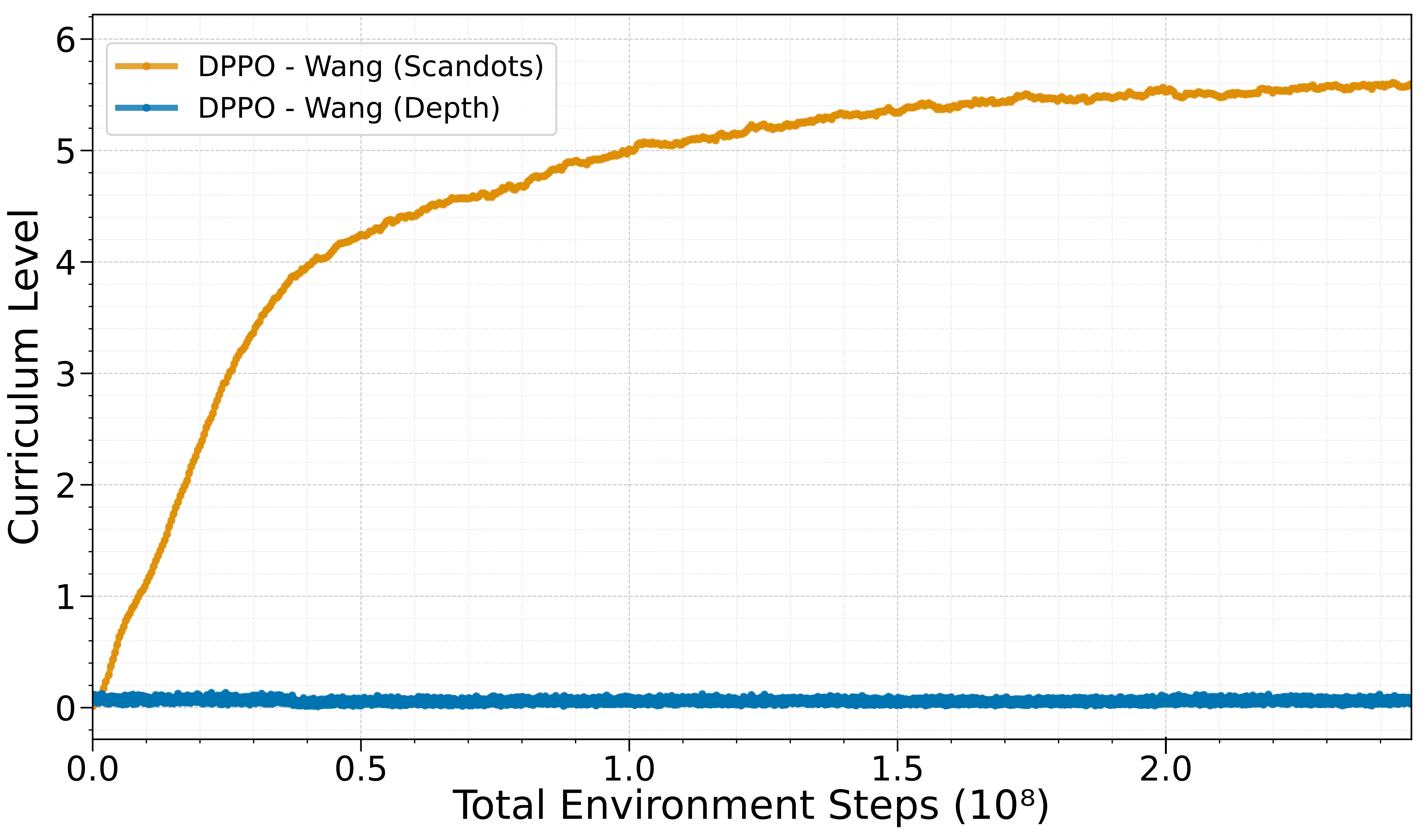}
  \caption{Navigation task curriculum progression. The privileged teacher (Scandots) makes stable progress, while the depth-based policy (Depth) fails to learn, likely due to noisy gradients from small batch sizes necessitated by the high VRAM cost of rendering many parallel high-dimensional depth images.}
  \vspace{-0.5cm}
  \label{fig:depth_plot_curriculum}
\end{figure}

To avoid the computational bottleneck of training directly on rendered depth images (Fig. \ref{fig:depth_plot_curriculum}), we first train a risk-aware teacher policy $\pi_{\theta}(a_{t}|o_{t}^{teacher},\beta)$ using low-dimensional privileged observations. The input \(o_{t}^{teacher}=(\tilde{h}_t, s_{t}^{robot}, s_{t}^{goal}, s_{t}^{task})\) includes a ground-truth height scan, robot state, and task goals (detailed in Table \ref{tab:obs_unified}).
Importantly, we condition the policy on the risk-sensitivity parameter \(\beta\), enabling online risk adaption during deployment.
Details of action spaces and reward functions can be found in Tables \ref{tab:actions} and \ref{tab:reward_unified}.
Privileged height scan information $\tilde{h}_t$ is first encoded through a small 2-layer CNN, then concatenated with the remaining observations $[s^{robot}_t, s^{goal}_t, s^{task}_t, \beta]$ and passed through an 1-layer LSTM followed by a 3-layer MLP to produce actions. 
We represent our critic using a QR-DQN \cite{dabney2018distributional}, which models the risk-neutral value distribution \(Z_{\phi}(s)\) as a uniform mixture of Dirac delta functions centred at state-dependent quantiles \(\phi_{i}(s)\) predicted by the critic network:  
\begin{equation}
    Z_{\phi}(s) = \frac{1}{N}\sum^{N}_{i=1}\delta_{\phi_{i}(s)},
    \label{eq: quantile_dist}
\end{equation}
where \(\delta_{z}\) denotes a Dirac distribution centred at \(z\in\mathbb{R}\), and \(\{\phi_{1}(s), \allowbreak\dots, \allowbreak\phi_{N}\}\) are the learned quantile values.
To introduce risk sensitivity to actor policy updates, we apply a distortion risk metric \cite{majumdar2020distortion} to the predicted return distribution \(Z_{\phi}(s)\) (as shown in Phase 1 of Fig. \ref{fig:method_diagram}), to obtain a distorted value distribution. Following \cite{Schneider-ICRA-24, ma2020dsac}, we compute the distorted expectation as a weighted sum over the distributions risk-adjusted quantile supports:
\begin{equation}
V_\beta(s) = \int_0^1 g_\beta^{\prime}(\tau) Z_\phi^\tau(s) \, d \tau=\sum_{k=1}^N\left(g_\beta\left(\tau_k\right)-g_\beta\left(\tau_{k-1}\right)\right) \phi_k(s),
\label{eq: distorted expectation}
\end{equation}

where \(\tau_i=\frac{i}{N}\) for \(i=0,\dots,N\) are the quantile fractions of the distribution, \(g_{\beta}(\tau)\) is a distortion function that modifies the quantile fractions based on the risk-sensitivity parameter \(\beta\), and \(Z_\phi^\tau(s)\) denotes the \(\tau\)-quantile of the distribution. Similar to \cite{Schneider-ICRA-24}, the risk-sensitive value estimate \(V_{\beta}(s)\) is used in advantage estimates for PPO policy updates.
In this work we investigate policies trained with the Wang \cite{wang2000class} and \ac{CVaR} \cite{rockafellar2000optimization} distortion risk metrics, whose distortion functions \(g_{\beta}(\tau)\) are defined as: 

\begin{equation}
g_\beta^{\text{Wang}}(\tau) = \Phi\left(\Phi^{-1}(\tau) + \beta\right),
\label{eq:wang_metric}
\end{equation}

\begin{equation}
g_{\beta}^{\text{CVaR}}(\tau) = \min\left\{\frac{\tau}{\beta}, 1\right\},
\label{eq:cvar_metric}
\end{equation}

where \(\Phi\) is the standard normal distribution, and \(\beta \in \mathbb{R}\) is the scalar risk sensitivity parameter that parametrises the distortion function. Varying $\beta$ reweighs low-return versus high-return quantiles when computing the distorted value used in policy updates, directly controlling risk-averse or risk-seeking behaviour rather than acting as a generic weight on a single reward term.
For the Wang metric, \(\beta=0\) produces an unchanged advantage estimate, and therefore a risk-neutral policy, while \(\beta>0\) and \(\beta<0\) produce risk-averse and risk-seeking advantage estimates respectively. In contrast, \ac{CVaR} yields a risk-neutral policy when \(\beta=1\), and becomes increasingly risk-averse as \(\beta \rightarrow 0\). The distortion metric re-weights probability mass in the distribution according to the risk parameter $\beta$: for example, $g^{Wang}_{\beta>0}(\tau)$ (risk-averse) pushes mass to the low-value tail, making bad outcomes appear more likely, penalising risky states.


\subsection{Risk-aware Visuomotor Student Policy (Phase 2)}
\label{Sec: method: student policy}

As privileged height scan information \(\tilde{h}_t\) is not available on the real robot, we need to distil the risk-aware teacher policy into a student policy that acts only on observations available from the real hardware. 
To do so, we use supervised learning to train a deployable student policy \(\pi_{\psi}\) conditioned on observations \(o_{t}^{\text{student}} = (d_t, s_{t}^{\text{robot}}, s_{t}^{\text{goal}}, s_{t}^{\text{task}})\), where \(d_t\) is an egocentric depth image, \(s_{t}^{\text{robot}}\) is the robot's proprioceptive state, \(s_{t}^{\text{goal}}\) is the task goal, and \(s_{t}^{\text{task}}\) contains non-privileged task-specific information.  
To accommodate this change, we replace the teacher's input height scan CNN encoder with a depth-image CNN encoder $E_{depth}(d_t)$ of matching output dimension, reusing the teacher's LSTM+MLP architecture (initialised with teacher weights). The student policy \(\pi_{\psi}\) is trained with DAgger \cite{ross2011reduction} to minimise the $L_2$ loss between student actions and teacher actions.
We first step the environment with teacher actions for 600 episodes to mitigate distributional shift, updating only the CNN depth encoder weights. We then unfreeze all student policy weights and step the environment with student actions. An overview of our method is provided in Fig. \ref{fig:method_diagram}.


\section{Experiments}


Experiments use the Toyota HSR \cite{ToyotaHSR} mobile manipulator, featuring a holonomic base and a 5-DoF arm. To simplify the state space, its head-mounted depth camera is fixed facing downward to capture the immediate frontal workspace. The learned policy operates at 30Hz, issuing velocity commands to the mobile base and joint position commands to the arm, resulting in an action space of $\mathcal{A} \subseteq \mathbb{R}^8$.  
Depth observation dimensions are $87\times58$, updated at 10Hz.
We train policies in IsaacLab \cite{mittal2023orbit} with the RSL-RL library \cite{Schneider-ICRA-24}, training risk-aware teacher policies using \ac{DPPO} \cite{Schneider-ICRA-24}. We apply our method on two tasks: navigation and object picking, which differ in action/observation spaces, reward structure and training curriculum. For the navigation task, teacher and student observation spaces are $\mathcal{O} \subseteq \mathbb{R}^{168}$ and $\mathcal{O} \subseteq \mathbb{R}^{5091}$, respectively. For the pick task, $\mathcal{O} \subseteq \mathbb{R}^{170}$ and $\mathcal{O} \subseteq \mathbb{R}^{5094}$. For all evaluation metrics and task performance plots, confidence bounds are calculated as in \cite{agarwal2021deep}.

\vspace{-0.1cm}

\begin{figure}[t]
    \centering
    
    \begin{subfigure}[t]{0.455\columnwidth}
        \centering
        \includegraphics[
            width=\linewidth, 
            trim={200pt 50pt 30pt 50pt}, 
            clip
        ]{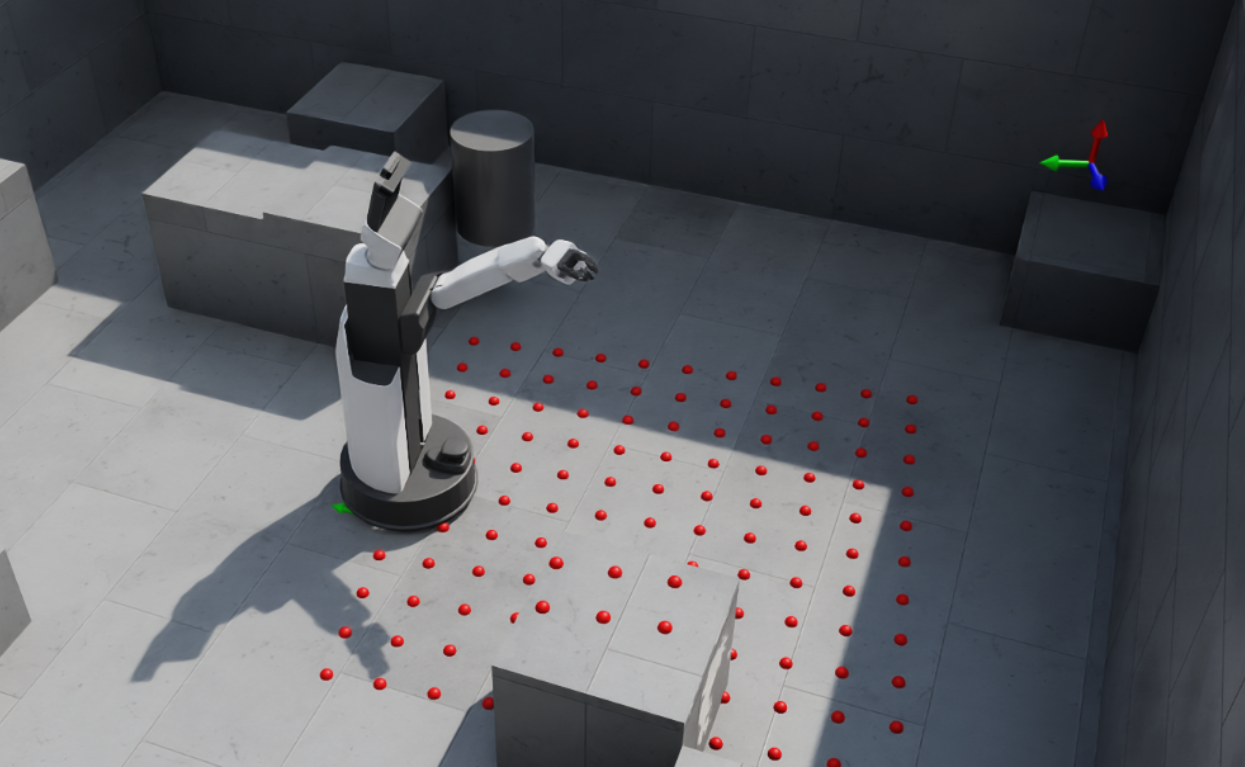}
        \caption{Navigation Task}
        \label{fig:nav_task_world}
    \end{subfigure}
    \hfill 
    \begin{subfigure}[t]{0.52\columnwidth}
        \centering
        \includegraphics[
            width=\linewidth,
            trim={30pt 50pt 180pt 50pt},
            clip
        ]{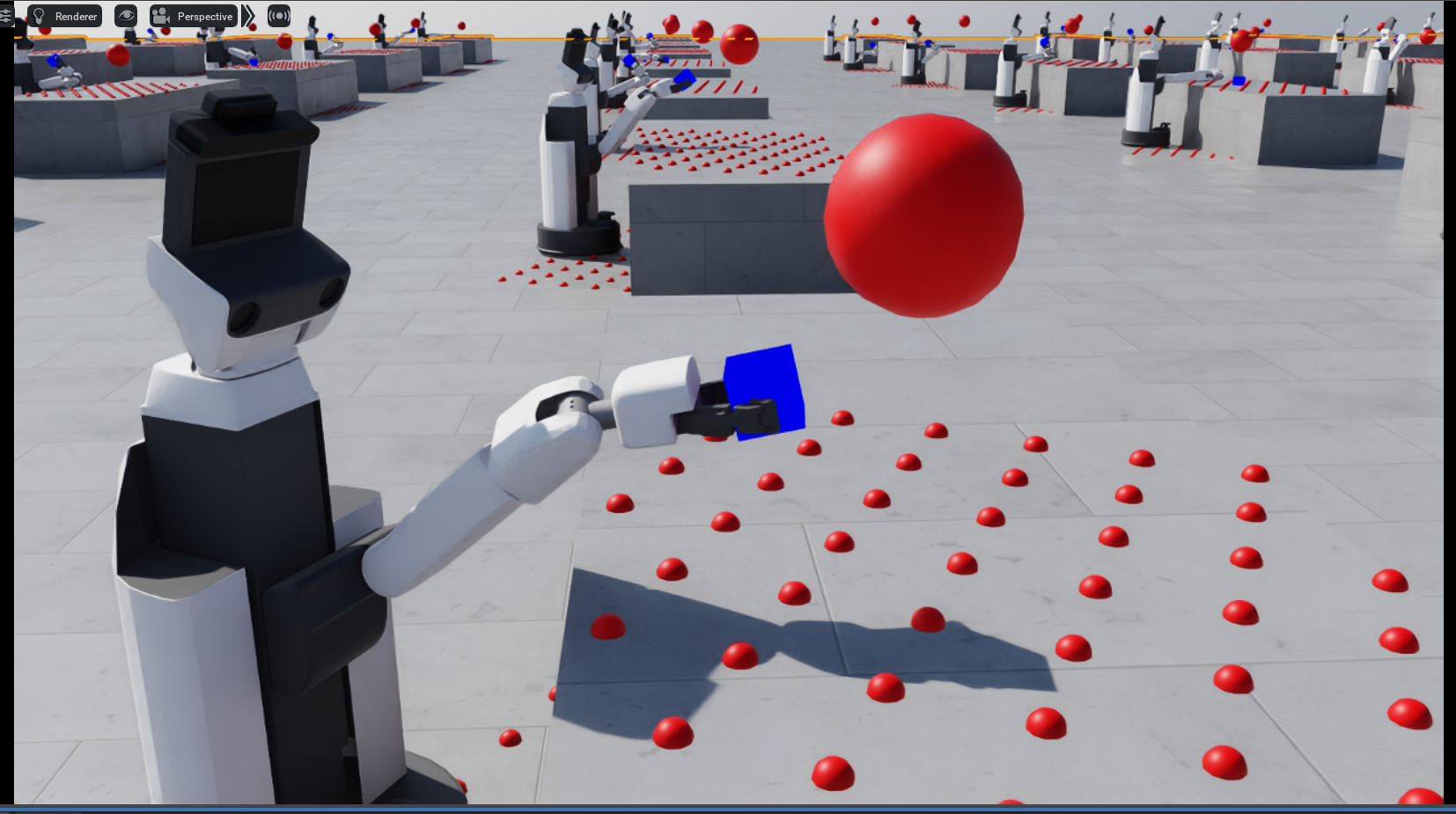}
        \caption{Pick Task}
        \label{fig:pick_task_world}
    \end{subfigure}

    \caption{Training environments. Left: Navigation task, reaching a 3D target (axes) while avoiding static and dynamic obstacles. Right: Pick task, grasping and lifting a cube to a goal (red sphere).}
    \label{fig:task_envs}
    \vspace{-0.5cm} 
\end{figure}

\textbf{Navigation Task:} The robot must perform whole-body control to reach a 3D target in a procedurally generated environment populated with static obstacles and a randomly moving cylinder (Fig. \ref{fig:nav_task_world}). This dynamic obstacle introduces aleatoric uncertainty into the transition dynamics. The policy must balance reaching the goal (which triggers successful termination) against the risk of collisions, which incur penalties and early termination. 

\textbf{Pick Task:} The robot must grasp and lift a cube to a target position above a table (Fig. \ref{fig:pick_task_world}). Episodes terminate with a penalty if the robot base collides with the environment or if the cube is displaced beyond a recoverable threshold (e.g., dropped or pushed too far). Unlike the navigation task, goal success does not terminate the episode; rather, rewards accumulate at every timestep the goal is maintained. This requires the policy to balance swift execution against the risk of aggressive, terminating actions.
Rewards (Table \ref{tab:reward_unified}) are adapted from ManiSkill3's PickCube-v1 task \cite{tao2024maniskill3}.

\begin{figure}[!ht]
    \centering

    \begin{subfigure}[t]{\columnwidth}
        \centering
        \includegraphics[width=\linewidth, clip]{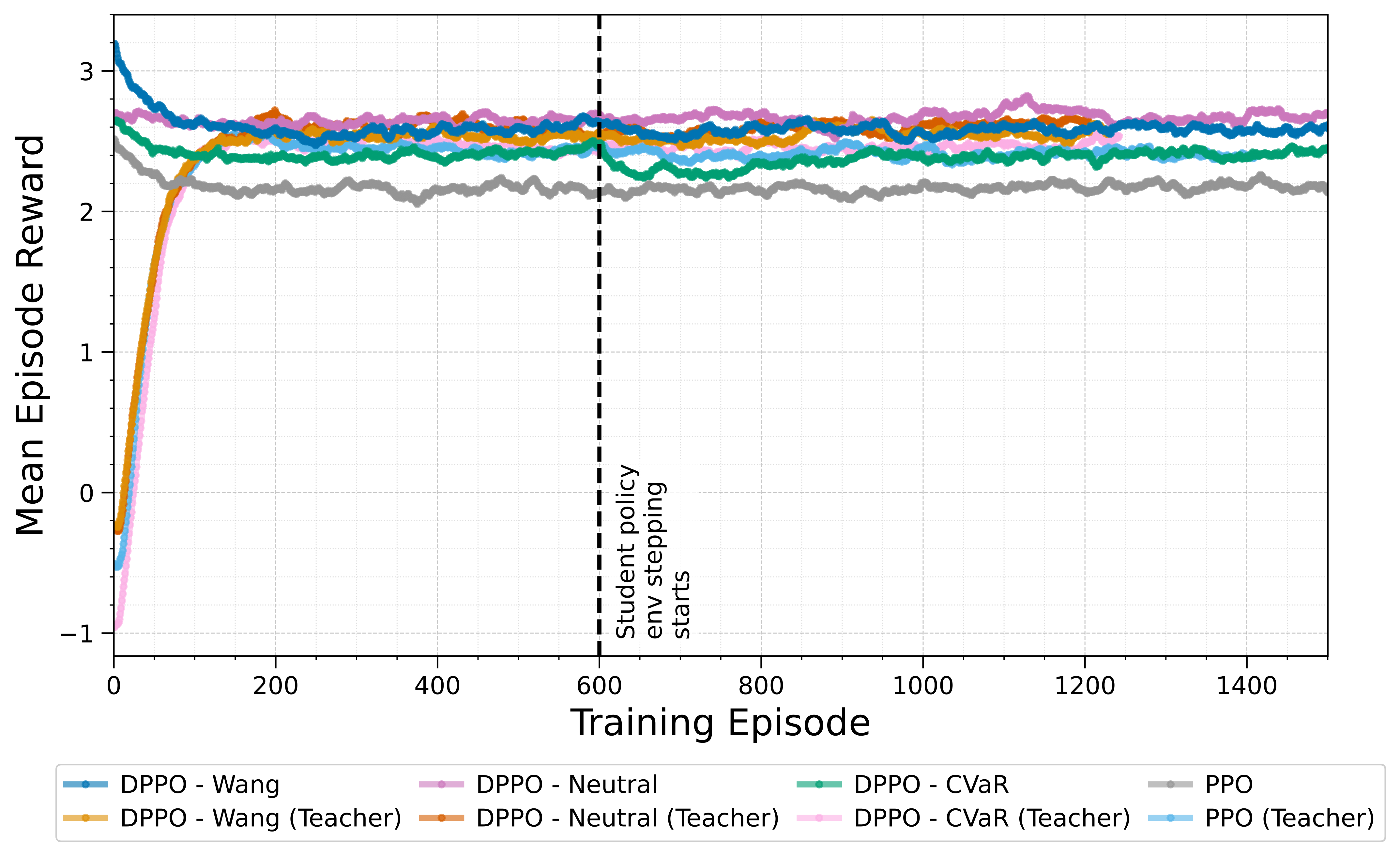}
        \caption{Navigation task.}
        \label{fig:nav_task_training_curves}
    \end{subfigure}

    \medskip 

    \begin{subfigure}[t]{\columnwidth}
        \centering
        \includegraphics[width=\linewidth, clip]{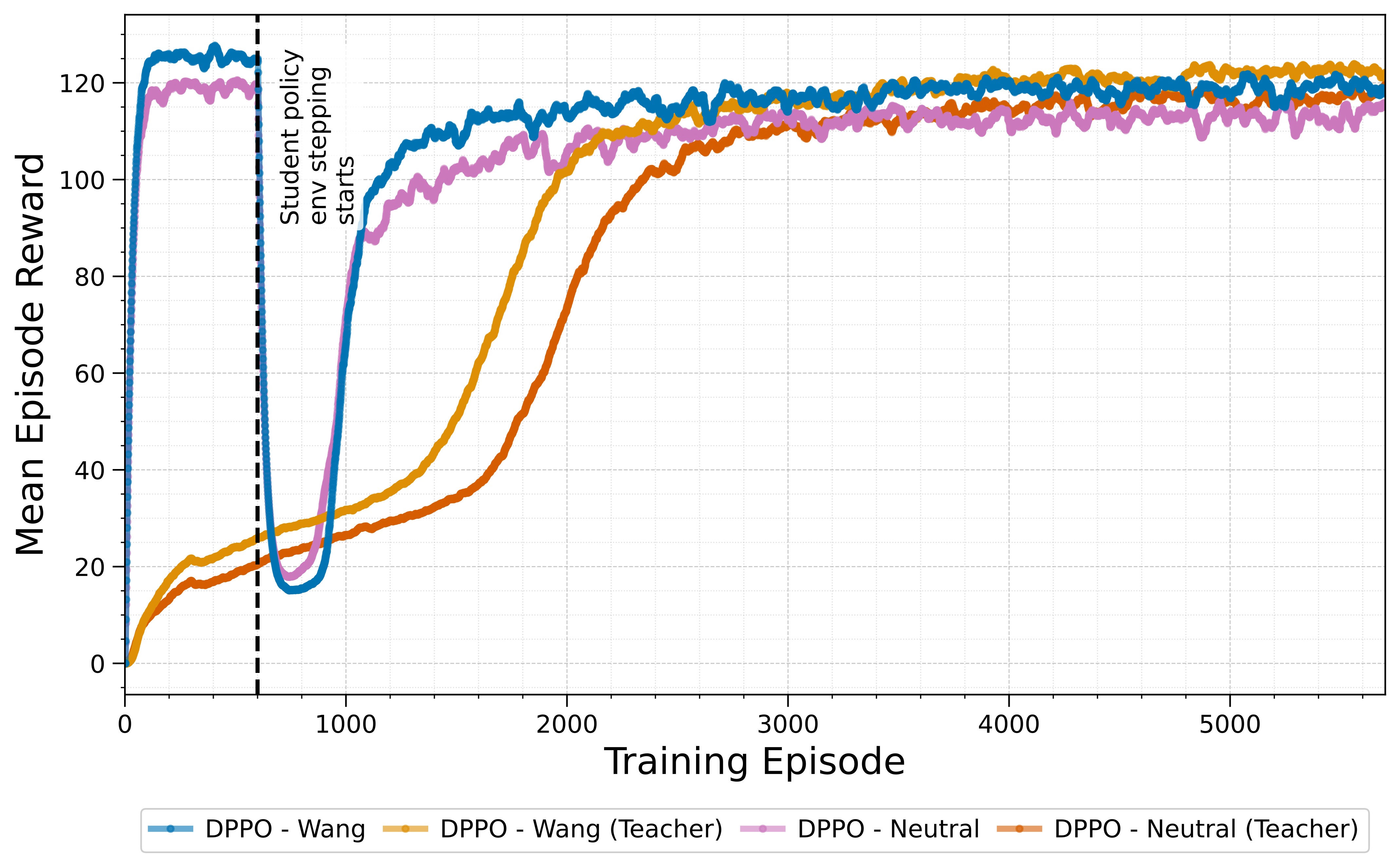}
        \caption{Pick task.}
        \label{fig:pick_task_training_curves}
    \end{subfigure}

    \caption{Training curves for (a) navigation and (b) pick tasks. Teachers evaluate \(16\times\) more steps per episode due to higher parallel environment counts (\(4096\) vs.\ \(256\)). The drop at episode 600 in (b) marks the switch to student-driven environment stepping.}
    \vspace{-0.5cm} 
\end{figure}

\textbf{Training:} 
For baselines we also train risk-neutral \ac{DPPO} and PPO \cite{schulman2017proximal} policies. Each task makes use of a manually designed curriculum (see Table \ref{tab:curricula_unified}) and is trained with domain randomisation \cite{tobin2017domain}. During student training, we set curriculum level progression to the maximum level for each task, and apply the same domain randomisation techniques as seen during the teacher policy training. 
During training we uniformly sample \(\beta\in[-1.0,1.0]\) for teacher policies using the Wang metric, and \(\beta\in[0.01,1.0]\) for those using \ac{CVaR}.
In the pick task, we successfully train a risk-aware \ac{DPPO} policy using the Wang metric, and a standard (risk-neutral) \ac{DPPO} policy. However, we are unable to train PPO or \ac{DPPO}-\ac{CVaR} policies for this task. We hypothesise this is due to superior performance of \ac{DRL} methods over their non-distributional counterparts \cite{bdr2023}, and the well-documented challenges in training \ac{CVaR} policies \cite{Schneider-ICRA-24}. Training curves for teacher and student policies are shown in Figs. \ref{fig:nav_task_training_curves} and \ref{fig:pick_task_training_curves}. Policies were trained using a NVIDIA RTX A6000, hyperparameters are provided in Table \ref{tab:hyperparams}.

\begin{table}[hb!]
\centering
\caption{Actions per task (applied at 30\,Hz).}
\label{tab:actions}
\begin{tabular}{@{}lcc@{}}
\toprule
\textbf{Action Component} & \textbf{Navigation} & \textbf{Pick} \\
\midrule
Base cmd. vel. $\mathbf{v}^{\text{base}}_t=(v_x, v_y, \omega_z)$ & \Yes & \Yes \\
Arm joint positions $\mathbf{q}_t$ & \Yes & \Yes \\
Gripper close/open $a_t^{\text{grip}}\in\{0,1\}$ & \Dash & \Yes \\
\bottomrule
\vspace{-0.6cm}
\end{tabular}
\end{table}

\begin{table}[ht]
\centering
\caption{Hyperparameters for RL training across navigation and pick tasks. The same hyperparameters were used across all baselines trained on a task. No extensive hyperparamter searches were performed. Bracketed values denote student distillation settings.}
\label{tab:hyperparams}
\UseUnifiedTableStyle
\begin{tabular}{lcc}
\toprule
\textbf{Hyperparameter} & \textbf{Navigation Task} & \textbf{Pick Task} \\
\midrule
Learning rate & $1 \times 10^{-3}$ & $1 \times 10^{-3}$ \\
Discount factor $\gamma$ & 0.99 & 0.99 \\
GAE($\lambda$), $\lambda$ & 0.95 & 0.95 \\
Batch size & 128 & 256 \\
Epochs per update & 5 & 5 \\
Clip parameter $\epsilon$ & 0.2 & 0.15 \\
Entropy coefficient & $-1\times 10^{-3}$ & $-5\times 10^{-4}$ \\
Value loss coefficient & 0.9 & 0.7 \\
Max gradient norm & 1.0 & 1.0 \\
Schedule & Adaptive & Adaptive \\
Target KL-divergence & 0.01 & 0.01 \\
Number of parallel environments & 4096 (256) & 4096 (256) \\
Number of steps per episode & 96 & 48 \\
\midrule
\multicolumn{3}{c}{\textbf{QR-DQN Critic Network Specific Hyperparameters}} \\
\midrule
Number of Quantiles & \multicolumn{2}{c}{200} \\
$SR(\lambda)$, \hspace{0.1cm} $\lambda$ & \multicolumn{2}{c}{0.95} \\ 
Value loss function & \multicolumn{2}{c}{Quantile L1} \\
\bottomrule
\vspace{-0.4cm}
\end{tabular}
\end{table}

\begin{table}[t]
\caption{Observations and noise terms. \Yes = included; \YesN = included with noise; \Dash \space = not included. TA (Teacher Actor) and S (Students) can receive noise; TC (Teacher Critics) have no additive noise (following \cite{dexterous_inhand_2020}). Transforms are in the robot base frame; base pose/velocity via wheel odometry.}
\label{tab:obs_unified}
\vspace{-2mm}
\centering
\UseUnifiedTableStyle
\setlength{\tabcolsep}{2pt}
\resizebox{\columnwidth}{!}{%
\begin{tabular}{@{}l c c ccc ccc@{}}
\toprule
\multirow{2}{*}{\textbf{Component}} & \multirow{2}{*}{\textbf{Noise}} & \multirow{2}{*}{\textbf{Params}} &
\multicolumn{3}{c}{\textbf{Navigation}} &
\multicolumn{3}{c}{\textbf{Pick}} \\
\cmidrule(lr){4-6}\cmidrule(lr){7-9}
& & & \textbf{TA} & \textbf{TC} & \textbf{S} & \textbf{TA} & \textbf{TC} & \textbf{S} \\
\midrule
\multicolumn{9}{@{}l}{\textit{High-level / exteroception}} \\
Egocentric depth $\mathbf{d_t}$                   & $\mathcal{U}(-a,a)$ & $a=0.1\,\mathrm{m}$ & \cdash & \Dash & \YesN & \Dash & \Dash & \YesN \\
Height scan (1D) $\mathbf{\tilde{h}_t}$           & \Dash               & \Dash               & \Yes  & \Yes  & \Dash & \Yes  & \Yes  & \Dash \\
Risk sensitivity $\mathbf{\beta}$                 & \Dash               & \Dash               & \Yes  & \Dash & \Yes  & \Yes  & \Dash & \Yes  \\
\midrule
\multicolumn{9}{@{}l}{\textit{Proprioceptive state} $s_t^{\text{robot}}$} \\
Base pose $\mathbf{x_t^{\text{base}}}$         & \Dash               & \Dash               & \Yes  & \Yes  & \Yes  & \Yes  & \Yes  & \Yes  \\
Base velocity $\mathbf{\dot{\mathbf{p}}_t^{base}}$   & \Dash               & \Dash               & \Yes  & \Yes  & \Yes  & \Yes  & \Yes  & \Yes  \\
Joint positions $\mathbf{q_t}$                      & $\mathcal{U}(-a,a)$ & $a=0.01$            & \YesN & \Yes  & \YesN & \YesN & \Yes  & \YesN \\
Joint velocities $\mathbf{\dot{\mathbf{q}}_t}$               & $\mathcal{U}(-a,a)$ & $a=0.01$            & \YesN & \Yes  & \YesN & \YesN & \Yes  & \YesN \\
EE pose $\mathbf{x_t^{\text{ee}}}$              & $\mathcal{U}(-a,a)$ & $a=0.01$            & \YesN & \Yes  & \YesN & \YesN & \Yes  & \YesN \\
Previous action $\mathbf{a_{t-1}}$                  & \Dash               & \Dash               & \Yes  & \Yes  & \Yes  & \Yes  & \Yes  & \Yes  \\
\midrule
\multicolumn{9}{@{}l}{\textit{Task observations} $s_t^{\text{task}}$} \\
EE target $\mathbf{g^{ee}}$                            & \Dash               & \Dash               & \Yes  & \Yes  & \Yes  & \Dash & \Dash & \Dash \\
EE--goal dist.\ $\mathbf{p_t^{ee}}-\mathbf{g^{\text{ee}}}$ & \Dash & \Dash     & \Yes  & \Yes  & \Yes  & \Dash & \Dash & \Dash \\
Masked cylinder pos. $\mathbf{\tilde{\mathbf{p}}_t^{cyl}}$ & $\mathcal{U}(-a,a)$ & $a=0.05\,\mathrm{m}$ & \YesN & \Dash & \Dash & \Dash & \Dash & \Dash \\
Cylinder pos. $\mathbf{{\mathbf{p}}_t^{cyl}}$   & \Dash               & \Dash               & \Dash & \Yes  & \Dash & \Dash & \Dash & \Dash \\
Object target $g^{\text{obj}}$                       & \Dash               & \Dash               & \Dash & \Dash & \Dash & \Yes  & \Yes  & \Yes  \\
Masked object $\mathbf{\tilde{\mathbf{p}}_t^{obj}}$   & $\mathcal{U}(-a,a)$ & $a=0.03\,\mathrm{m}$ & \Dash & \Dash & \Dash & \YesN & \Yes  & \YesN \vspace{0.05cm} \\
\vspace{0.05cm}Object visible $\mathbb{I_{\textbf{t}}^{\textbf{visible}}}$      & \Dash               & \Dash               & \Dash & \Dash & \Dash & \Yes  & \Yes  & \Yes  \\
\vspace{0.05cm}Grasp success $\mathbb{I_{\textbf{t}}^{\textbf{grasp}}}$         & \Dash               & \Dash               & \Dash & \Dash & \Dash & \Yes  & \Yes  & \Dash \\
EE--object dist.\ $\|\mathbf{{p_t^{ee}}-\mathbf{p}_t^{obj}}\|$ & \Dash & \Dash & \Dash & \Dash & \Dash & \Yes & \Yes & \Yes \\
Obj.--goal dist.\ $\|\mathbf{p_t^{obj}}-\mathbf{g^{obj}}\|$          & \Dash & \Dash & \Dash & \Dash & \Dash & \Yes & \Yes & \Yes \\
\bottomrule
\end{tabular}
}
\vspace{-5mm}
\end{table}



\begin{table*}[th]
\centering
\caption{Curricula for navigation and pick teacher policies. Rows cover (i) reward weight updates and (ii) level/difficulty progression, including triggers and parameter ranges.}
\label{tab:curricula_unified}
\scriptsize
\resizebox{\textwidth}{!}{%
\begin{tabular}{@{}llp{5.8cm}p{5.6cm}p{2.1cm}@{}}
\toprule
\textbf{Task} & \textbf{Aspect} & \textbf{Item} & \textbf{Setting / Details} & \textbf{Trigger / When} \\
\midrule
\multirow{7}{*}{\textbf{Navigation}} 
& \multirow{4}{*}{Reward schedule} 
& Action rate $\|\mathbf{a}_t - \mathbf{a}_{t-1}\|$ & Update weight $w_i = -1\times 10^{-4}$ & Iter. 400 \\
& & Arm joint velocity $\|\dot{\mathbf{q}}^{\text{arm}}_t\|$ & Update weight $w_i = -2.5\times 10^{-3}$ & Iter. 400 \\
& & Base velocity $\|\mathbf{v}^{\text{base}}_t\|$ & Update weight $w_i = -5\times 10^{-3}$ & Iter. 400 \\
& & EE acceleration $\|\dot{\mathbf{v}}^{\text{ee}}_t\|$ & Update weight $w_i = -2.5\times 10^{-3}$ & Iter. 800 \\
\cmidrule(l){2-5}
& \multirow{3}{*}{Level progression} 
& Levels & 10 total; tracked per environment & — \\
& & Level change rules & Success: $+1$ (cap at max; at max sample levels uniformly). Failure/termination: $-1$ (min 0). & On episode outcome \\
& & Goal $(x,y)$ range & Increases with level: $\pm0.25\,$m $\rightarrow\,\pm2.0\,$m & — \\
\midrule
\multirow{7}{*}{\textbf{Pick}} 
& \multirow{1}{*}{Reward schedule} 
& Arm joint velocity $\|\dot{\mathbf{q}}^{\text{arm}}_t\|$ & Update weight $w_i = -0.1$ & Iter. 300 \\
& & Action rate $\|\mathbf{a}_t - \mathbf{a}_{t-1}\|$ & Update weight $w_i = -1$ & Iter. 3500 \\
\cmidrule(l){2-5}
& \multirow{6}{*}{Difficulty / initial configuration} 
& Levels & 20 total; tracked per environment & — \\
& & Level change rules & Successful grasp: $+1$ (cap at max; at max sample levels uniformly). No grasp/termination: $-1$ (min 0). & On grasp/termination \\
& & Initial EE–object distance & Sampled with level: $[0.0,\,0.5]\,$m & Harder with $\uparrow$ level \\
& & Object $(x,y)$ position perturbation & Around nominal: $[-0.2,\,0.2]\,$m & Harder with $\uparrow$ level \\
& & Joint configuration perturbation & Around nominal: $[-0.5, 0.5]\,$rad & Harder with $\uparrow$ level \\
\bottomrule
\end{tabular}
}
\end{table*}

\begin{table}[t]
\caption{Reward terms formulas and weights.
$r_t=\Delta t\sum_i w_i r_t^{(i)}$, $\Delta t=\text{simulation dt}=8.33{\times}10^{-3}$. Terms $\lVert \mathbf{F}_{L,R}\rVert$, $\mathbf{F}_{\min}$, $d_{L,R}$, $\theta_{\max}$ represent absolute contact force in left/right gripper fingers, threshold contact force $(0.5N)$, the open directions of the gripper fingers, and minimum angle between contact force and open direction $(85^{\circ})$ for a valid grasp. \(T\) is the number of timesteps per episode. }
\label{tab:reward_unified}
\vspace{-1mm}
\centering
\UseUnifiedTableStyle
\setlength{\tabcolsep}{1.5pt}
\resizebox{\columnwidth}{!}{%
\begin{tabular}{@{}%
  >{\raggedright\arraybackslash}p{.25\columnwidth}%
  >{\centering\arraybackslash}p{.485\columnwidth}%
  >{\raggedleft\arraybackslash}p{.1\columnwidth}%
  >{\raggedleft\arraybackslash}p{.1\columnwidth}@{}}
\toprule
\textbf{Component / Term} & \textbf{Formula} & \multicolumn{2}{c}{\textbf{Weights $w_i$}} \\
\cmidrule(lr){3-4}
& & \multicolumn{1}{c}{Navigation} & \multicolumn{1}{c}{Pick} \\
\midrule
EE goal & 
$15\cdot\,\mathbb{I}\big(\lVert \mathbf{x_t^{ee}}-\mathbf{g^{ee}}\rVert<0.15\,\mathrm{m}\big)$
& 10 & --- \\

EE shaped & 
$\lVert \mathbf{p_t^{ee}}-\mathbf{g^{ee}}\rVert-\lVert \mathbf{p_{t-1}^{ee}}-\mathbf{g^{ee}}\rVert$
& 10 & --- \\

EE--object & 
$1-\tanh\!\big(5\cdot\,\lVert \mathbf{p}_t^{ee}-\mathbf{p}_t^{obj}\rVert\big)$
& --- & 1.0 \\

Grasp success $\mathbb{I}_t^{\text{grasp}}$ &
\parbox{\linewidth}{\centering
  $\displaystyle \mathbb{I}\Big(\ \lVert \mathbf{F}_{L,R}\rVert>\mathbf{F}_{\min}\ \wedge\
  \angle(\mathbf{F}_{L,R},\mathbf{d}_{L,R})\le \mathbf{\theta}_{\max}\ \Big)$
}
& --- & 5.0 \\

Obj.--goal & 
$1-\tanh\!\big(\lVert \mathbf{p}_t^{obj}-\mathbf{g^{goal}}\rVert\big)$
& --- & 5.0 \\

$\mathbb{I}_{\mathrm{placed}}(s_t,\rho):=$ &
$\mathbb{I}\big(\,\lVert \mathbf{p}_t^{obj}-\mathbf{g}^{goal}\rVert<\rho\,\big)$
& --- & --- \\

$\mathbb{I}_{\mathrm{vel}}(s_t,\nu):=$ &
$\mathbb{I}\big(\,\lVert \mathbf{v}_t^{obj}\rVert<\nu\,\big)$
& --- & --- \\

Static vel.\ & 
$\bigl(1-\tanh\bigl(\tfrac{\lVert\dot{\mathbf{q}}_t^{\mathrm{arm}}\rVert+\lVert\mathbf{v}_t^{\mathrm{base}}\rVert}{3}\bigr)\bigr)\cdot
\mathbb{I}_{\mathrm{placed}}(s_t,0.25)$
& {\text{---}} & 10.0 \\

Pick success & 
$\mathbb{I}_{\mathrm{vel}}(s_t,1.5)\cdot\mathbb{I}_{\mathrm{placed}}(s_t,0.15)$
& --- & 20.0 \\

Termination \ $\mathbb{I}_t^{\text{termination}}$ &
$\mathbb{I}\big(\text{terminal at }t\big)$
& $-5.0$ & $-20.0$ \\

Alive\ $\mathbb{I}_t^{\text{alive}}$ &
$\mathbb{I}\big(\text{episode not terminal at }t\big)$
& $-0.03$ & --- \\
Padded alive\ & 
$\mathbb{I}_t^{\text{termination}}\,\cdot(T - t)$
& $-0.03$ & --- \\

Action rate  & $\lVert\mathbf{a}_t-\mathbf{a}_{t-1}\rVert$ & $-5{\times}10^{-5}$ & $-1{\times}10^{-4}$ \\
Arm vel.\ & $\lVert\dot{\mathbf{q}}_t^{arm}\rVert$ & $-2.5{\times}10^{-4}$ & $-0.03$ \\
Arm acc.\ & $\lVert\ddot{\mathbf{q}}_t^{arm}\rVert$ & $-2.5{\times}10^{-5}$ & --- \\
Base vel.\ & $\lVert\mathbf{v}_t^{base}\rVert$ & $-5{\times}10^{-4}$ & $-0.01$ \\
Base acc.\ & $\lVert\dot{\mathbf{v}}_t^{base}\rVert$ & $-1{\times}10^{-4}$ & --- \\
EE acc.\ & $\lVert\dot{\mathbf{v}}_t^{ee}\rVert$ & $-5{\times}10^{-4}$ & --- \\
\bottomrule
\end{tabular}
}
\vspace{-6mm}
\end{table}


To evaluate the aggregate performance of our trained policies, we perform extensive experiments across both tasks. 
For the navigation task, we generate a held-out test set of 32 environments, and evaluate both our teacher and student policies by running 25 rollouts in each environment, resulting in 800 evaluation rollouts. For our risk-aware policies, we repeat this evaluation across a range of risk-sensitivity parameters. 
For the pick task, we create 32 distinct environments with varying table heights. For each environment, we create a test set of 25 randomly-sampled initial object locations and corresponding goal locations, again repeating across a range of risk-sensitivities. We also investigate risk-aware behaviour on the pick task by performing 800 rollouts in a challenging single instance of the environment, with the cube placed deep on the table near the edge of the robots reachable workspace. 
Policies trained with the Wang metric are evaluated at \(\beta=[-1.0,-0.5, 0.0, 0.5,1.0]\), and \ac{CVaR} policies are evaluated at \(\beta=[0.05, 0.15, 0.25, 0.5, 1.0]\). We found that performing \ac{CVaR} policy rollouts with \(\beta < 0.05\) led to severe performance degradation. We hypothesise this deterioration likely arises as low values of \(\beta\) emphasise a very small portion of the estimated value distribution, resulting in noisy and unreliable risk-adjusted value estimates.


\section{Results}

\begin{figure*}[t]
    \centering
    
    \begin{subfigure}[t]{\linewidth}
        \centering
        \includegraphics[
            width=\linewidth, 
            clip
        ]{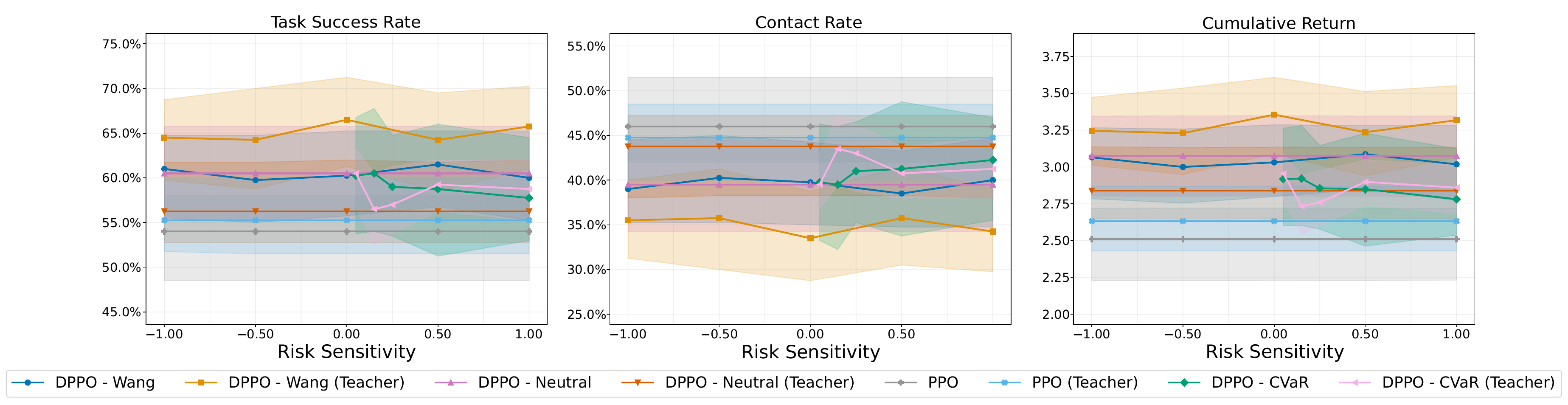}
        \caption{Navigation Task}
        \label{fig:nav_task_rates}
    \end{subfigure}
    \hfill 
    \begin{subfigure}[t]{\linewidth}
        \centering
        \includegraphics[
            width=\linewidth,
            clip
        ]{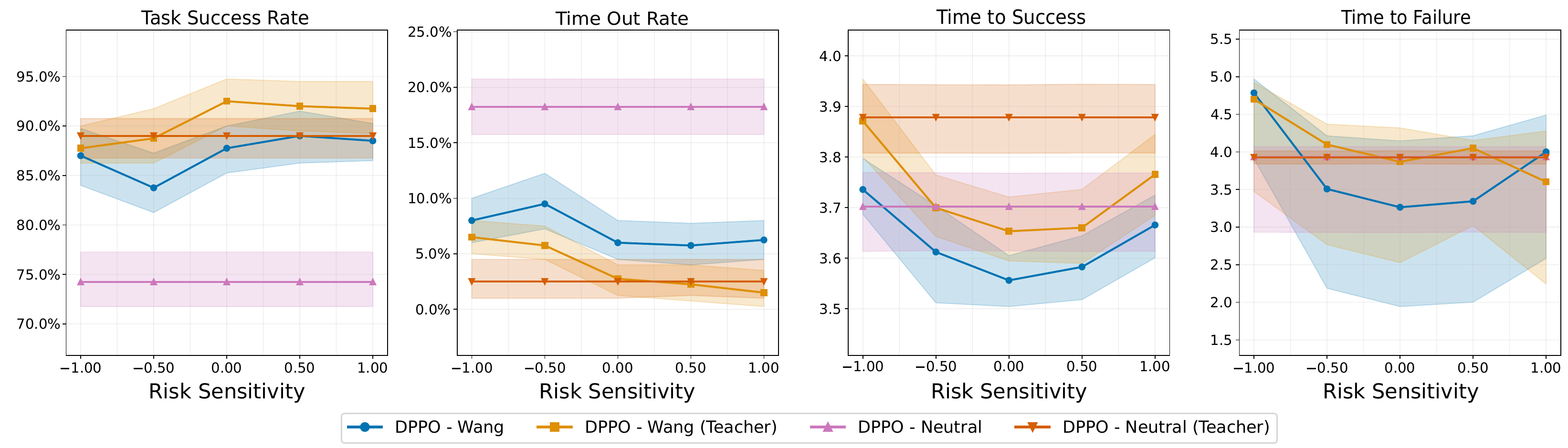}
        \caption{Pick Task}
        \label{fig:pick_task_rates}
    \end{subfigure}

    \caption{Task rates for policies evaluated on the navigation and pick tasks. (a) Left to right: task success rate; contact rate with the environment and the dynamic obstacle; cumulative return. (b) Left to right: task success rate; time out rate (not termination); time to reach goal success; time to task failure termination.}
    \label{fig:task_envs}
    \vspace{-0.5cm} 
\end{figure*}

\begin{figure*}[h]
  \centering
  \includegraphics[
            width=0.995\linewidth, 
            clip
        ]{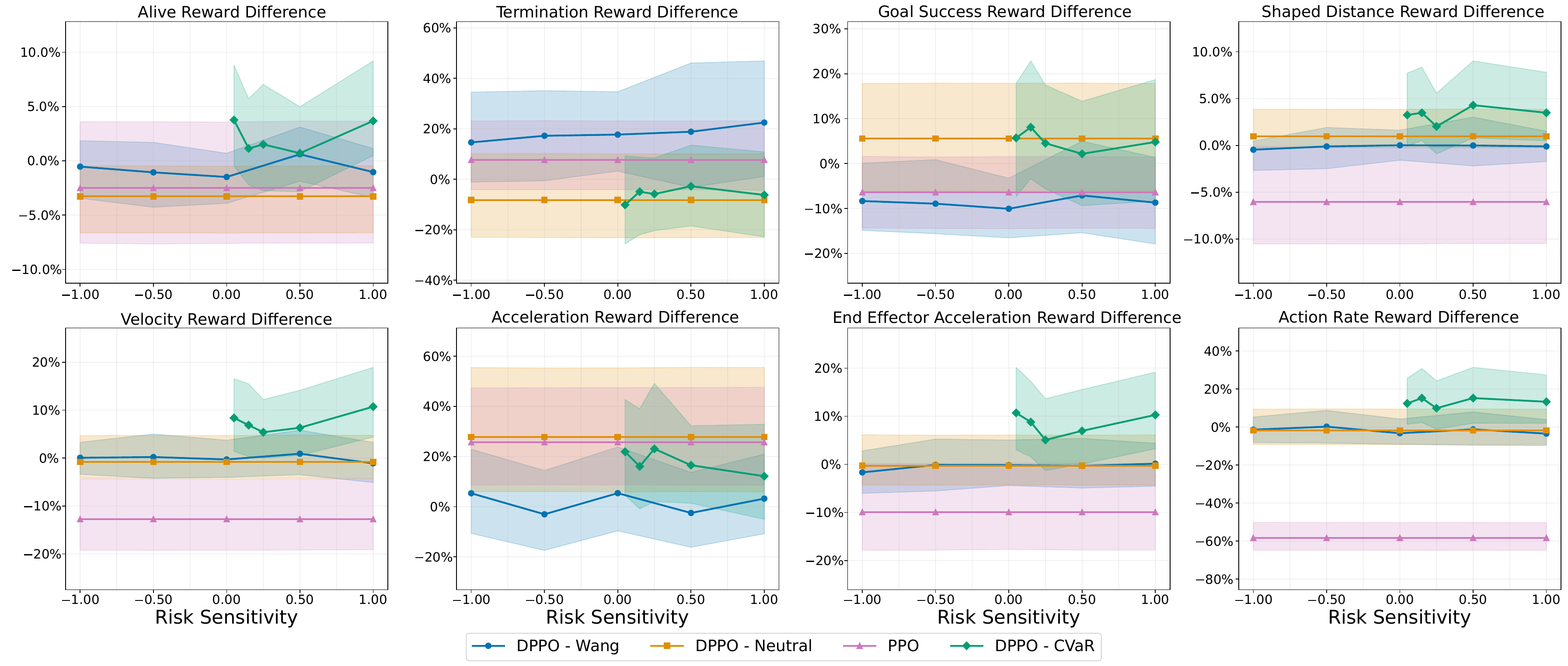}
  \caption{Difference in cumulative individual reward terms during the navigation task evaluation.}
  \vspace{-0.4cm}
  \label{fig:nav_task_rewards_difference}
\end{figure*}

\begin{figure}[htbp]
  \centering
  \includegraphics[width=\columnwidth,  trim=0.0cm 0.0cm 0.0cm 0.0cm, clip]{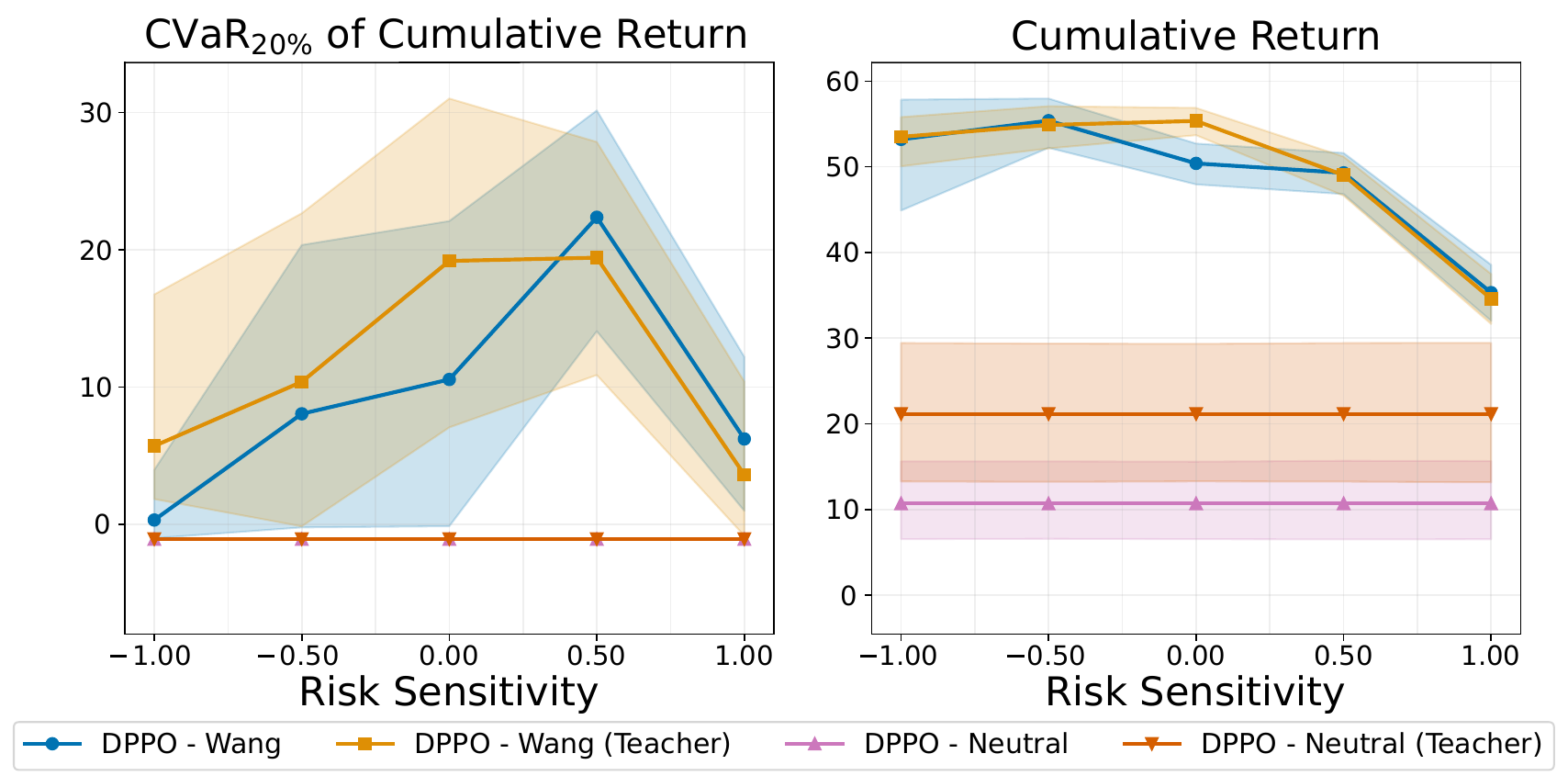}
  \caption{Plots of policies evaluated on a single instance of the pick task. Right: Cumulative return. Left: \(20\%\) \ac{CVaR} of cumulative return. \(20\%\) is used as a balanced trade off between statistical reliability on the limited number of considered rollouts while placing emphasis on worst-case outcomes.}
  \vspace{-0.8cm}
  \label{fig:pick_task_cvar}
\end{figure}



A key aspect of our approach is the transfer of learned risk-aware behaviours from teacher to student policies. We investigate this by comparing the \textit{difference} in teacher-student rewards in sampled navigation task environment-goal pairs, plotting the individual reward terms  used in the navigation task (Fig. \ref{fig:nav_task_rewards_difference}). 
Ideally, successful transfer manifests as flat lines in these plots, indicating a constant performance gap in teacher-student, regardless of risk-sensitivity. As shown in Fig. \ref{fig:nav_task_rewards_difference} student policies in the navigation task maintain this stable difference, suggesting our approach effectively transfers learnt risk-aware behaviour to student policies. 
Delving deeper into individual reward components in the navigation task (Fig. \ref{fig:nav_task_rewards_difference}), we see that reward terms with higher weightings, which heavily influence the behaviour of learnt teacher policies, are transferred well to students while terms with lower weightings (e.g., acceleration-based penalties) show larger relative difference. This aligns with expectations; the teacher's training maximises risk-adjusted return which is dominated by high-weight rewards. Behaviour driven by these terms is more prevalent and thus transfers more readily during \ac{IL}. 

To explicitly quantify risk-aware behaviour, we measure catastrophic events via collision and timeout rates. Task performance metrics for both the pick and navigation tasks are plotted in Figs. \ref{fig:nav_task_rates} and \ref{fig:pick_task_rates}, where risk-aware student policies achieve overall performance comparable to risk-neutral baselines. 
These results demonstrate, for the first time, a framework capable of learning effective, risk-aware visuomotor policies that achieve competent behaviour, establishing the viability of our approach.
Interestingly, while DPPO-Neutral and DPPO-Wang evaluated at $\beta=0$ might be expected to perform identically, DPPO-Wang is trained on trajectories induced by a mixture of $\beta$ values. This changes the state distribution encountered during training, leading to differences in the learned policy even when evaluated at a risk-neutral $\beta=0$. Exploring whether multi-$\beta$ training improves overall performance is left for future work. Related works have similarly leveraged uncertainty and risk estimates to improve \ac{RL} training \cite{rigter2023risk, burda2018exploration}. 
Given the complexity of the multi-reward tasks we train on, emergent risk-aware behaviours can often remain hidden beneath competing reward signals. To investigate any risk-aware behaviour produced by policies trained with our framework, we plot the \(20\%\) \ac{CVaR} of cumulative return from the evaluations performed on the single instance of the pick task, shown in Fig. \ref{fig:pick_task_cvar}. This figure shows the average return of the \(20\%\) worst rollouts, illustrating the policies worst-case performance. In this figure we would expect risk-averse policies \((\beta>0)\) to achieve better worst-case performance, which we demonstrate in Fig. \ref{fig:pick_task_cvar}, indicating our visuomotor student policies show risk-aware behaviours. Conversely, Fig. \ref{fig:pick_task_cvar} shows that risk-seeking policies achieve a higher average cumulative return, though with more variability in performance (indicated by wider estimated confidence bounds). Qualitatively, this can be explained by risk-seeking policies attempting grasps sooner, meaning more time can be spent near the goal state and therefore higher return is achieved. However, this approach can sometimes result in poor quality grasps explaining the variability in performance.  


\section{Conclusion}
\label{sec:conclusion}
In this paper we present a framework to train risk-aware visuomotor policies to reduce severe outcomes caused by aleatoric uncertainty. 
We propose a two-phase framework, first training risk-aware teacher policies on privileged observations, then transferring these capabilities to policies conditioned on depth images\linebreak[3] via \acs{IL}.
Our approach trains risk-aware visuomotor policies for mobile manipulation conditioned on egocentric depth observations with runtime-adjustable risk sensitivity, achieving comparable performance to risk-neutral methods, and we show learnt risk-aware behaviours can be successfully transferred through \ac{IL}. 
This work provides a practical pathway for deploying risk-aware controllers for mobile manipulators using rich, high-dimensional sensor data, enabling more dependable behaviour in complex, dynamic environments.

\section{Limitations and Future Work}

While our framework successfully demonstrates the transfer of risk-aware behaviours to visuomotor policies, we note several limitations that provide exciting avenues for future research:

\textbf{Hardware Validation:} Our evaluations are conducted entirely in simulation. Deploying our risk-aware visuomotor policies to physical mobile manipulators to validate sim-to-real transfer is a critical next step. \newline
\textbf{Uncertainty Modelling:} By trusting the critic's value distribution, we are implicitly assuming pure aleatoric uncertainty. Incorporating risk-aversion to epistemic uncertainty is vital for handling real-world distributional shifts. \newline
\textbf{Risk-Aware Student Objective:} Currently student policies are trained solely via an \ac{IL} objective. Future work should fine-tune bootstrapped student policies directly on a risk-aware \ac{RL} objective. \newline
\textbf{Reward Shaping:} Relying on shaped rewards for training stability risks incorrect task specification \cite{booth2023perils}. Exploring intrinsic motivation and reward methods for sparse-reward settings is an important direction. \newline
\textbf{Extreme Risk Sensitivities:} We observe a trend where policies evaluated at the absolute bound of our trained risk-sensitivities $(\beta=\pm1)$ exhibit degraded risk-aware behaviour. Investigating training paradigms that stabilise performance across wider and more extreme ranges of $\beta$ remains an open challenge. \newline
\textbf{Task Complexity:} The tasks evaluated in this work occur in relatively simple simulated environments. Scaling our approach to handle visually complex, cluttered, and realistic scenes will be necessary to fully test the limits of our risk-aware visuomotor policies.





\vspace{-0.25cm}
\bibliographystyle{IEEEtran}
\bibliography{IEEEtranBST/IEEEabrv,IEEEtranBST/IEEEexample}

\end{document}